\pdfoutput=1

\documentclass[11pt]{article}

\usepackage[]{ACL2023}

\usepackage{times}
\usepackage{latexsym}
\usepackage[T1]{fontenc}
\usepackage{booktabs} 
\usepackage{multirow, makecell}
\usepackage{tabularx}
\usepackage{bbding}
\usepackage{pifont}
\usepackage{graphicx}
\usepackage{array,ragged2e}

\usepackage[utf8]{inputenc}
\usepackage{microtype}
\usepackage{inconsolata}

\newcommand{\datasetname}{INLI }

\newcommand{\COMMENT}[1]{}

\title{Entailed Between the Lines: Incorporating Implication into NLI}

\author{Shreya Havaldar$^{*\diamond}$, Hamidreza Alvari$^\dag$, John Palowitch$^\dag$, \\ \bf Mohammad Javad Hosseini$^\dag$, Senaka Buthpitiya$^\dag$, Alex Fabrikant$^\dag$ \\
$^\diamond$University of Pennsylvania, $^\dag$Google Deepmind \\
\texttt{shreyah@seas.upenn.edu, \{hamidrz, palowitch\}@google.com} \\}

\begin{document}
\maketitle
\begin{abstract}

\begingroup\def\thefootnote{*}\footnotetext{Research done during internship at Google Deepmind}\endgroup

Much of human communication depends on implication, conveying meaning beyond literal words to express a wider range of thoughts, intentions, and feelings. For models to better understand and facilitate human communication, they must be responsive to the text's implicit meaning. We focus on Natural Language Inference (NLI), a core tool for many language tasks, and find that state-of-the-art NLI models and datasets struggle to recognize a range of cases where entailment is implied, rather than explicit from the text. We formalize \textbf{implied entailment} as an extension of the NLI task and introduce the Implied NLI dataset (INLI) to help today's LLMs both recognize a broader variety of implied entailments and to distinguish between implicit and explicit entailment.\footnote{Data \& resources available at \url{https://github.com/google-deepmind/inli}} We show how LLMs fine-tuned on INLI understand implied entailment and can generalize this understanding across datasets and domains.
\end{abstract}

\section{Introduction}

Human communication is rich with implication. Emotions, social cues, insults, and a myriad of other messages are conveyed implicitly, often even more so than explicitly. In order for LLMs to reach human-level understanding of communication, they must be able to understand a text's implications.

\begin{figure}[t]
    \centering
    \includegraphics[width=0.45\textwidth]{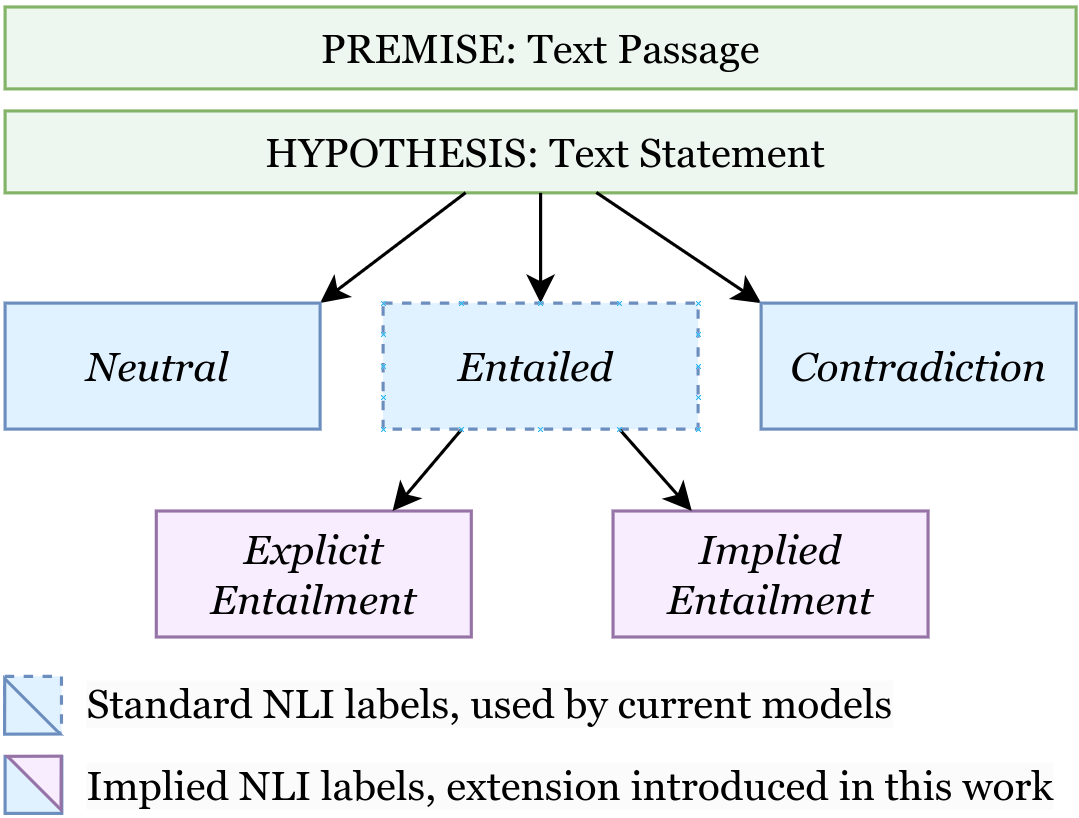}
    \caption{Extending NLI to introduce implication. We propose that entailment models should treat implications as entailed, while also distinguishing between explicit and implicit entailments.}
    \label{fig:concept}
\end{figure}

\newcommand{\xmark}{\ding{55}}%
\begin{table*}[t]
\small
  \centering
  \begin{tabular}{p{0.3\textwidth}p{0.25\textwidth}ccc}
    \toprule
    \textbf{Premise} & \textbf{Hypothesis} & \textbf{Entailed} & \textbf{Implicit} &\textbf{Label}\\
    \midrule
    \multirow{1}{=}{Alex walked towards the checkout lane in the grocery store. The cashier asked if he had brought a bag for his groceries. With an angry grumble, Alex cursed himself for forgetting and told the clerk he needed to purchase a bag that evening.} & Alex needs to purchase a bag for his groceries. & \Checkmark & \xmark & Explicit entailment\\ 
    & The grocery store does not provide free bags. & \Checkmark & \Checkmark & Implied entailment\\ 
    & The cashier at the checkout lane is middle-aged. & \xmark & -- & Neutral\\ 
    & Alex remembered to bring a bag to the grocery store. & \xmark & -- & Contradiction\\ 
    \bottomrule 
  \end{tabular}
  \caption{Examples of the four labels in our implied entailment task. Unlike the standard entailment task, we draw the distinction between explicitly entailed and implicitly entailed hypotheses. Implications must be \textit{inferable} (i.e. fully entailed in the premise) and \textit{implicit} (i.e. not explicitly stated in the premise). Note that the definitions of neutral and contradictory remain the same as the standard task.}
  \label{tab:examples}
\end{table*}

Consider what a human reader learns from the following sentence: \textit{After reading the ARR reviews, Kim had to go stuff himself with cheesecake.} 

Beyond the explicit text that informs us that (a) Kim read ARR reviews, and (b) Kim subsequently was obliged to eat cheesecake, we also readily absorb a variety of implicitly entailed facts:  (c) Kim found reading the reviews unpleasant, (d) Kim actually proceeded to eat cheesecake, and (e) Kim ate more than an ordinary amount of cheesecake. 

Modern LLMs take on a number of roles and applications where understanding \textit{implied entailments} like (c-e) is crucial. For instance, in the context of creative writing or mental health, it is vital to consider (c) to understand emotional states, often communicated implicitly \cite{gullestad2019theory}. In the context of translation or cross-cultural communication, where patterns of implicit language use vary across languages and cultures, \cite{kim1998high, havaldar2023comparingstyleslanguages}, it is important to understand the subtlety between (d) and (b), and to translate the figurative meaning of (e) rather than the verbatim hyperbole.


Prior work on implied language in NLP predominantly focuses on rigorously structured inputs -- indirect answers to yes/no questions \cite{george2020conversational, damgaard-etal-2021-ill, louis2020d}, pairwise entity selection \cite{hosseini2023resolvingindirectreferringexpressions}, and logical consistency \cite{jeretic2020natural, zheng2021grice}. While useful for pragmatic understanding, these tasks require inputs that have a specific structure, and thus have limited applicability when it comes to LLM reasoning on natural, unstructured premises. 

NLI models (i.e. classifiers to decide if a hypothesis $H$ is entailed by, contradicted by, or neutral with respect to a premise $P$) have been used for a variety of NLP applications -- grounding, factuality, proposition segmentation, etc. \cite{stowe-etal-2022-impli, bhagavatula2020abductivecommonsensereasoning, gao2023rarrresearchingrevisinglanguage, hosseini2024aps}. However, as we show in this paper, standard NLI datasets contain few implications. As a result, models trained on these benchmarks struggle to recognize implied entailments as entailments at all. Ensuring these cases are covered by entailment classifiers is important if LLMs are to understand nuances and subtleties in human communication. 

Additionally, several applications of NLI, such as summarization evaluation or citation generation, would significantly benefit from a \textit{distinction} between explicit and implied entailment, thus motivating our dataset's 4-way refinement to separate these entailment labels within the traditional 3-way NLI taxonomy.

In this paper, we introduce the \textbf{implied entailment} task and build a dataset on which NLI models can be fine-tuned to recognize implications as valid entailments. Concretely, when determining whether a given premise $P$ entails a hypothesis $H$, we want NLI models to reason about whether $H$ is \textit{implicitly} or \textit{explicitly} entailed in $P$ (Figure \ref{fig:concept}). Specifically, our contributions are as follows:

\begin{enumerate}
\vspace{-0.5em}
\itemsep 0em
    \item We \textit{formalize the implied entailment task}, distinguishing implied entailment from explicit entailment and introducing an NLI paradigm that reflects this distinction.
    \item We build an \textit{INLI -- the implied NLI dataset}, with 10k premises mirroring real-world communication and 40k hypotheses that are implied, explicit, neutral, and contradictory.
    \item We show that NLI models fine-tuned on INLI are able to recognize when a hypothesis is implied versus explicitly entailed, and that they \textit{generalize across domains and datasets.}

\end{enumerate}

\section{Formalizing Implied Entailment} 
\label{sec:definition}

We start with the conventional 3-way distinction of NLI labels \cite{williams-etal-2018-broad}: Given a premise $P$ and a hypothesis $H$, $H$ is an entailed by $P$ if it is definitely correct given $P$, contradicted by $P$ if it is definitely incorrect given $P$, and neutral otherwise. 

We then further refine entailment into two categories: explicit and implied.
An implied entailment is a hypothesis that requires the reader to make additional cognitive deductions beyond the explicit language\footnote{Note that our definitions conflate the distinctions drawn in pragmatics/philosophy of literature between ``explicatures''  \cite{sperber1986relevance}, ``implicatures'' \cite{grice1991studies}, and ``entailments'' \cite{moldovan2019can}. In our framework, explicit entailment is equivalent to explicatures, while implied entailment combines implicatures and non-explicature entailments.} used in the text:



\begin{itemize}
\vspace{-0.5em}
\itemsep 0em
    \item \textbf{Explicit entailment: } Follows directly from the text's lexical semantics (e.g. via synonymy and paraphrasing) and syntax (e.g. via pronominal co-reference, bridging, or other endophora).
    \item \textbf{Implied entailment:} Requires some sort of an additional cognitive step, such as logical reasoning, world knowledge, conversational pragmatics, or figurative language.
\end{itemize}

We use these definitions to create an evaluation scheme for human annotation of INLI in Section~\ref{sec:annotation}. Table~\ref{tab:examples} shows examples of hypotheses in our 4-way implied entailment task.

\COMMENT{\subsection{Implied Entailment as an NLI Task}
The standard NLI task encourages models to learn the boundary between entailed vs. non-entailed hypotheses. Given the prevalence of implications in communication, we propose that it is also useful for models to learn the boundary between implicit and explicit entailments, a distinction that has not yet been formally incorporated into the NLI task. 
}


\section{Implied Language Understanding in Current Benchmarks \& Models}

To motivate the need for an implied entailment dataset, we first explore whether current NLI benchmarks already contain implicitly entailed hypotheses, and whether current NLI models already possess the ability to generalize to implications. Concretely, we answer the following questions:

\begin{enumerate}
\vspace{-0.5em}
\itemsep 0em
    \item To what extent do widely-used NLI benchmarks contain implied entailments?
    \item How accurately can current NLI models infer implied entailments?
\end{enumerate}

\subsection{NLI Benchmarks Contain Few Implied Entailments}
\label{sec:implicitness_model}

\begin{table}
\small
    \centering
    \begin{tabular}{lr}
    \toprule
    \textbf{Dataset}  & \textbf{\% Implied}\\
    \midrule
    SNLI \cite{bowman2015large} & 9.33\\
    MNLI \cite{williams-etal-2018-broad} & 3.68\\
    ANLI \cite{nie2020adversarialnlinewbenchmark} & 15.66\\
    WANLI \cite{Liu2022WANLIWA} & 5.48\\
    \bottomrule
    \end{tabular}
    \caption{Quantifying implied entailments present in existing NLI benchmarks. Most NLI benchmarks contain a disproportionate amount of explicit entailments, highlighting the need for an implication-focused dataset.}
    \label{tab:curr-benchmarks}
\end{table}

INLI (see Section~\ref{sec:dataset}) contains both implicit and explicit entailments. Using this data, we fine-tune a \texttt{T5-XXL} model \cite{raffel2020exploring} to distinguish between implicit and explicit entailments, achieving 97.3\% test accuracy. We then apply this model to entailments in existing NLI benchmarks -- the Stanford NLI Corpus \cite{bowman2015large}, Multi-Genre NLI \cite{williams-etal-2018-broad}, Adversarial NLI \cite{nie2020adversarialnlinewbenchmark}, and Worker-and-AI NLI \cite{Liu2022WANLIWA}.

Table~\ref{tab:curr-benchmarks} shows what percent of each of these benchmarks are classified as implied entailments by our fine-tuned model  -- most contain very few implied entailments, suggesting that the vast majority of entailed hypotheses in current benchmarks are explicit. The notable exception is ANLI, a dataset designed to adversarially distill exemplars that stump models. ANLI contains three subsets, or ``rounds,'' with increasing difficulty, and we observe that the number of implied entailments increases as the rounds get harder. 

This further suggests that modern NLI models struggle with understanding implication, as the hypotheses that challenge models most are implicit in nature. See Appendix~\ref{app:experiments-curr-benchmarks} for details on experiment setup and model training.

\paragraph{Human validation of Table~\ref{tab:curr-benchmarks}.} Training on implicit and explicit entailments from INLI and then testing on existing benchmarks requires manual validation to ensure our implicitness detection model can properly generalize. We verify that annotators agree with 92.0\% of model outputs, with an inter-annotator agreement (Cohen's Kappa) of 0.768. Appendix~\ref{app:experiments-curr-benchmarks} contains additional details about annotation setup and results.

\subsection{NLI Models Rarely Generalize to Implied Entailments}

\begin{table}[t]
\small
    \centering
    \begin{tabular}{lrr}
    \toprule
     & \multicolumn{2}{c}{\textbf{Entailment Accuracy}} \\
    \cline{2-3} \rule{0pt}{1.25em}
    \textbf{Training Dataset}  & Implied & Explicit\\
    \midrule
    SNLI & 0.500 & 0.943 \\
    MNLI & 0.528 & 0.965 \\
    ANLI & 0.714 & 0.983 \\
    WANLI & 0.525 & 0.905 \\
    
    \bottomrule
    \end{tabular}
    \caption{Measuring how well models fine-tuned on existing NLI datasets can infer explicit entailments and implied entailments in INLI. Results suggest existing NLI datasets do not support generalization to implied entailments.}
    \label{tab:curr-models}
\end{table}

\renewcommand{\arraystretch}{1.25}
\begin{table*}[t]
    \centering
    \small
    \begin{tabular}{p{2.5cm}p{3.5cm}p{8.5cm}}
    \toprule
    \textbf{Dataset} & \textbf{Implicature Frame} & \textbf{Example} \\
    \midrule 
    
     \textsc{Ludwig} \cite{george2020conversational} & \texttt{Question} \newline 
     \texttt{Indirect Answer} \newline
     \texttt{Implied Meaning} &
     \texttt{Question:} Would you like to go to a party tonight? \newline
     \texttt{Answer:} I am too tired. \newline
     \texttt{Meaning:} No \\
     
     \textsc{Circa} \newline \cite{louis2020d} & 
     \texttt{Conversational Context} \newline
     \texttt{Question} \newline 
     \texttt{Indirect Answer} \newline
     \texttt{Implied Meaning} & 
     \texttt{Context:} Colleagues leaving work together on a Friday. \newline
     \texttt{Question:} Do you want to hang out later? \newline
     \texttt{Answer:} I could do with a stiff drink. \newline
     \texttt{Meaning:} Yes \\
     
     \textsc{NormBank} \newline \cite{ziems2023normbank} & \texttt{Behavior} \newline 
     \texttt{Situational Context} \newline
     \texttt{Implied Social Norm} & 
     \texttt{Behavior:} Play with your food \newline
     \texttt{Context:} Have a food fight in a restaurant setting \newline
     \texttt{Social Norm:} Taboo \\
 
     \textsc{SocialChem} \newline \cite{forbes2020social} & \texttt{Social Situation} \newline 
     \texttt{Implied Rule-of-Thumb} & 
     \texttt{Situation:} Telling my sister I would not donate a kidney to her. \newline
     \texttt{Rule-of-Thumb:} You shouldn't expect someone to donate their organ to you. \\

    \bottomrule
    \end{tabular}
    \caption{Datasets chosen for this work and their corresponding implicature frames. To ensure the implied entailments in \datasetname are of the highest quality, we transform existing implicature frames (i.e. data that contains an implicature) into the $\langle$premise, implied entailment$\rangle$ format required for NLI.}
    \label{tab:implicature_frames}
\end{table*}

\begin{figure*}[t]
    \centering
    \includegraphics[width=\textwidth]{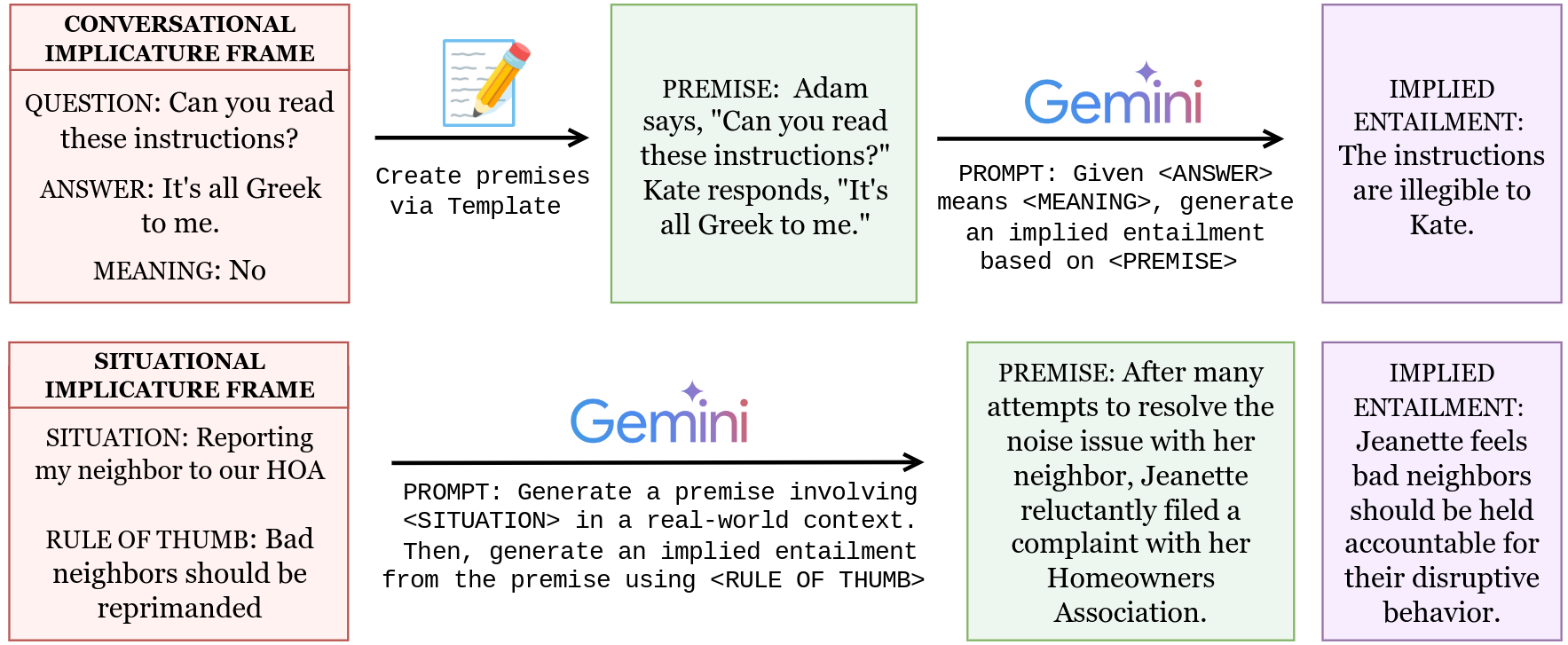}
    \caption{Stage 1: Implicature Augmentation. Given existing data that contains an implicature (e.g. the answer to an indirect question, a widely-accepted social norm, etc.), we augment the data into a premise and implicitly entailed hypothesis. In this figure we show augmented examples from \textsc{Ludwig} and \textsc{SocialChem}.}
    \label{fig:implicature_generation}
\end{figure*}

\begin{figure*}[t]
    \centering
    \includegraphics[width=\textwidth]{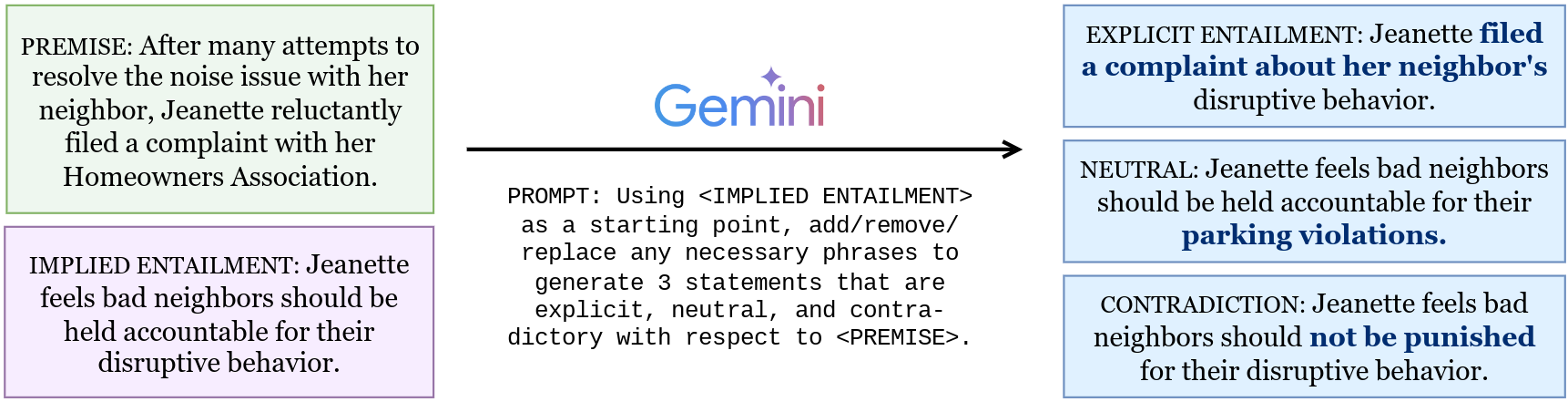}
    \caption{Stage 2: Alternative Hypothesis Generation. Given premises and their corresponding implied entailments, we generate three additional hypotheses (explicit, neutral, and contradictory) by subtly modifying the implied entailment. This ensures we build a more challenging dataset for the implied entailment task.}
    \label{fig:statement_generation}
\end{figure*}

\def\arraystretch{1.2}%
\begin{table*}[t]
  \centering
  \footnotesize
  \begin{tabular}{p{0.2\textwidth}p{0.7\textwidth}}
    \toprule

     \multicolumn{2}{l}{\textsc{\textbf{Ludwig} [n = 1,956]}} \\
     \midrule 
     \textit{Premise} & Diane says, "Would you like to go to a party tonight?" Sophie responds, "I am too tired."	\\ 
     \textit{Implied} & Sophie would prefer not to attend the party this evening.	\\ 
     \textit{Explicit} & Sophie claims to be too tired.	\\
     \textit{Neutral} & The party will take place outside.	\\
     \textit{Contradiction} & Sophie is excited to attend the party this evening. \\
     \midrule 
     \multicolumn{2}{l}{\textsc{\textbf{Circa} [n = 18,044]}} \\
     \midrule
     \textit{Premise} & Teresa and Alicia are colleagues who are leaving work on a Friday at the same time. Teresa says, ``Do you want to hang out later?'' Alicia responds, ``I could do with a stiff drink.''\\ 
     \textit{Implied} & Alicia hopes to spend time with Teresa later.	 \\ 
     \textit{Explicit} & Alicia could do with a stiff drink.	 \\
     \textit{Neutral} & Alicia would benefit from a night out.\\
     \textit{Contradiction} & Alicia doesn't feel like spending time with Teresa later. \\
     
     \midrule

      \multicolumn{2}{l}{\textsc{\textbf{NormBank} [n = 10,000]}} \\
     \midrule
       \textit{Premise} & A hush fell over the elegant restaurant as a group of diners, their laughter escalating hysterically, began to fling mashed potatoes and bread rolls at each other, splattering the pristine white tablecloth and surrounding patrons with food.\\ 
     \textit{Implied} & The other diners were shocked by the inappropriate food fight.		 \\ 
     \textit{Explicit} & The restaurant fell silent as a group of diners started throwing food at one another. \\
     \textit{Neutral} & The diners at the restaurant had never witnessed such a big food fight.	 \\
     \textit{Contradiction} & The other diners were entertained by the food fight and enthusiastically joined in.	 \\
     \midrule
      
      \multicolumn{2}{l}{\textsc{\textbf{SocialChem} [n = 10,000]}} \\
      \midrule
      \textit{Premise} & With a heavy heart, Maya explained to her sister, Chloe, that she wouldn't be able to donate a kidney, bracing herself for the wave of emotions that were sure to follow. \\ 
     \textit{Implied} & Maya realized her choice not to donate a kidney would probably devastate Chloe. \\ 
     \textit{Explicit} & Maya prepared herself for Chloe's emotional reaction when she told her she couldn't be a kidney donor. \\
     \textit{Neutral} & Maya realized her decision to move to another state would probably upset Chloe. \\
     \textit{Contradiction} & Maya felt confident that Chloe would be understanding of her choice not to donate her kidney. \\
     
    \bottomrule 
  \end{tabular}
  \caption{Examples from INLI. We augment implicature frames from four datasets to create premises and implicitly entailed hypotheses (see Table~\ref{tab:implicature_frames} for original data) and then use \texttt{Gemini-Pro} to generate the remaining alternative hypotheses. \textsc{[n]} indicates the number of $\langle$premise, hypothesis$\rangle$ pairs in INLI from each dataset.}
  \label{tab:dataset-examples}
\end{table*}

Despite the lack of implied entailments contained in NLI benchmarks, we also explore whether models trained on current datasets can successfully generalize and understand implications in INLI. 

We fine-tune a \texttt{T5-XXL} model on each of the four NLI benchmarks, and then measure their accuracy in classifying implicitly and explicitly entailed hypotheses in INLI. Table~\ref{tab:curr-models} contains these results. Unsurprisingly, we see very high accuracy in inferring explicit entailments, but for implied entailments, most models perform around 50\%, i.e. randomly guessing whether implied entailments are inferable or not.

The clear exception is the model trained on ANLI, suggesting that implied entailments in training data improves ability to infer implied entailments in other settings. However, there is still a significant performance gap between explicit vs. implied. Overall, Table~\ref{tab:curr-models} suggests a significant need for improvement in this task; see Appendix~\ref{app:experiments-curr-models} for experiment details.

\vspace{0.5em}
\noindent \textbf{Takeaways:} NLI benchmarks predominantly contain explicit entailments; consequently, NLI models struggle to infer implied entailments. These findings highlight the need for a dataset that focuses on \textit{implied entailment} across a variety of domains.

\section{The Implied NLI Dataset (INLI)}
\label{sec:dataset}

To construct INLI, we introduce a novel procedure to leverage existing datasets that exemplify language pragmatics. 

Specifically, given a structured example of implied meaning in a social context (see Table~\ref{tab:implicature_frames}), we prompt \texttt{Gemini-Pro} to derive a $\langle$premise, implied entailment$\rangle$ pair that can be used to train NLI models. This strategy is more efficient than traditional human-generation strategies of traditional NLI datasets \cite{bowman2015large, nie2020adversarialnlinewbenchmark} in the following three areas:

\begin{enumerate}
\vspace{-0.5em}
\itemsep 0em
    \item \textbf{Generation cost:} We sidestep the need to train annotators to generate implied hypotheses from premises.
    \item \textbf{Quality:} We derive implied entailments from pre-existing, human-vetted datasets, offering a high-quality foundation for INLI.
    \item \textbf{Reproducibility:} Our procedure is easily reproducible for researchers who want to expand or generalize INLI.
\end{enumerate}

Our pipeline to build INLI has two main components: (1) Augmenting Implicature Frames, and (2) Alternative Hypothesis Generation. We discuss each component in detail below.

\subsection{Augmenting Implicature Frames}

Given understanding implication is a complex and subjective task, we utilize past datasets that contain \textit{implicature frames}: structured templates that express an implied meaning arising from a conversational or behavioral setting. Table~\ref{tab:implicature_frames} gives examples of the four datasets we use in this work and their corresponding implicature frames. Our goal is to convert such implicature frames into $\langle$premise, hypothesis$\rangle$ pairs that can be used to train entailment models, where the hypothesis expresses the implied meaning present in the original frame. We select 10k frames from these datasets and few-shot prompt \texttt{Gemini-Pro} to perform this conversion. 

Specifically, we design an augmentation framework that takes in an implicature frame, like those in Table~\ref{tab:implicature_frames}, and extracts a corresponding premise and implied entailment. This process is different for conversational implicatures (\textsc{Ludwig, Circa}) and situational implicatures (\textsc{NormBank, SocialChem}) respectively.

\paragraph{Conversational implicatures.} For LLMs to properly grasp the nuances of communication, they must understand implications arising in conversations. Both \textsc{Ludwig} and \textsc{Circa} include questions and indirect answers. To construct a premise, we use a template with two random names from the 2011 US Census, simulating a back-and-forth conversation. We then prompt \texttt{Gemini-Pro} with this premise and the given yes/no implied meaning of the indirect answer, asking \texttt{Gemini-Pro} to generate an implied hypothesis based on the provided meaning (See top of Figure~\ref{fig:implicature_generation}).

\paragraph{Situational implicatures.} In order to excel in applications like creative writing or mental health, LLMs must also understand implications hinging on social norms and interpersonal relationships. For \textsc{NormBank} and \textsc{SocialChem}, we start by using the provided behavior and context, or social situation, to create a premise that reflects the given real-world scenario. Then, we prompt \texttt{Gemini-Pro} to generate an implied entailment from the premise based on the corresponding social norm or rule-of-thumb (See bottom of Figure~\ref{fig:implicature_generation}).

Table~\ref{tab:dataset-examples} contains the premises and implied entailments derived by augmenting the examples in Table~\ref{tab:implicature_frames}. Additional details on prompting procedure are provided in Appendix~\ref{app:prompts}.

\subsection{Alternative Hypothesis Generation}
After generating premises and implied entailments by augmenting implicature frames, we also want to generate the remaining hypotheses: explicit entailments, neutrals, and contradictions.

We experiment with various prompting techniques to accomplish this, analyzing results and iterating accordingly. Details on various prompting iterations are discussed in Appendix~\ref{app:prompting}, and our final prompt is in Appendix~\ref{app:prompts}.

In order for the hypotheses to be semantically similar (e.g. contradictions shouldn't have disproportionate negation) and make for a challenging NLI task, we have \texttt{Gemini-Pro} use the implied entailment as a starting point, and then replace any necessary words or phrases in order to transform the implied entailment into an alternative hypothesis that is explicit/neutral/contradictory in relation to the given premise. Figure~\ref{fig:statement_generation} details this process.

\paragraph{Paraphrasing.} As a final step, we have \texttt{Gemini-Pro} paraphrase all generated hypotheses to minimize artifacts in the data that could distinguish between hypothesis labels \cite{krishna2024paraphrasing, sadasivan2024aigeneratedtextreliablydetected}. We do not paraphrase the explicit entailments in our conversational implicature datasets, as these hypotheses are simpler and paraphrasing them might alter their true labels.

\paragraph{Data robustness.} Past research stresses that inference in NLI datasets should depend on both the premise and hypothesis \cite{poliak2018hypothesisbaselinesnaturallanguage}. A method to test premise-dependence is fine-tuning a model to label hypotheses without using the premises. Our dataset's hypothesis-only accuracy is similar to other benchmarks, suggesting minimal artifacts that dilute the difficulty of our task. Details in Appendix~\ref{app:robustness}.

\subsection{Human Validation of Dataset}
\label{sec:annotation}

To ensure that our pipeline yields high-quality data, all six authors of this paper annotate a subset of INLI. The annotation task is similar to that of previous NLI benchmarks -- annotators are only shown a premise and a hypothesis, and are instructed to label the hypothesis with either implied, explicit, neutral, or contradiction.

To define these four labels, annotators are given the definition in Section~\ref{sec:definition}. They are instructed to take into account \textit{inferability} and \textit{implicitness} of a hypothesis when assigning a label. 

\paragraph{Training human annotators.} The annotators were first given the annotation instructions (see Appendix~\ref{app:annotation}) and a subset of 16 $\langle$premise, hypothesis$\rangle$ calibration examples from INLI spanning all datasets and label types. The annotators then rated the calibration examples, and afterward were encouraged to discuss their answers with the other annotators, referring to the instructions to guide the discussion. This process was meant to allow annotators to refine their understanding of the implied entailment labels before proceeding into the task.

\paragraph{Annotation task.} For the final annotation task, we select 200 $\langle$premise, hypothesis$\rangle$ pairs from INLI (equally distributed across datasets and hypothesis types), and assign each annotator to label a different subset of 100; ultimately, we have three annotators labeling each example.

Annotators have a Fleiss's kappa $\kappa = 0.711$ and a majority agreement (i.e. at least 2 out of 3 annotators agree) with the INLI labels of 0.935. Our annotator agreement is in line with that of past NLI benchmarks (ANLI: $\kappa = 0.679 \mathrm{\;to\;} 0.721$, WANLI: $\kappa = 0.60$). Additionally, we see very high majority agreement with the INLI labels, indicating INLI examples reflect human intuition well.

\section{Learning Implied Entailment}

We now show the promise of INLI to imbue NLI models with understanding of implied entailment, and explore whether INLI leads to a generalizable understanding of implication. 
Specifically, we answer the following research questions:
\begin{itemize}
\vspace{-0.5em}
\itemsep 0em
    \item \textbf{RQ1:} Do models fine-tuned and few-shot prompted on INLI effectively reason about implied entailment?
    \item \textbf{RQ2:} Do models fine-tuned on INLI maintain efficacy on traditional NLI benchmarks?
    \item \textbf{RQ3:} Do models fine-tuned on INLI generalize across domains and datasets?
\end{itemize}

\begin{table}[t]
\newcolumntype{R}[1]{>{\RaggedLeft\arraybackslash}p{#1}}
\small
    \centering
    \begin{tabular}{lrr}
    \toprule
    \multirow{2}{*}{\textbf{Model}} & \multicolumn{2}{c}{\textbf{Accuracy}} \\
    \cline{2-3} \rule{0pt}{1.25em}
    & \textbf{Overall} & \textbf{Implied Entailment} \\
    \midrule
    \multicolumn{3}{l}{\textit{Fine-tuned LLMs}} \\
    \midrule
     \texttt{T5-Small} & 0.813 & 0.731 \\
     \texttt{T5-Base} & 0.871 & 0.817 \\
     \texttt{T5-Large} & 0.913 & 0.870 \\
     \texttt{T5-XXL} & 0.924 & 0.885 \\
    \midrule
    \multicolumn{3}{l}{\textit{8-shot Prompted LLMs}} \\
    \midrule
    \texttt{GPT-4o} & 0.749 & 0.608 \\
     \texttt{GPT-4} & 0.753 & 0.645 \\
     \texttt{Claude-3-Sonnet} & 0.686 & 0.738 \\
     \texttt{Mistral-Large} & 0.744 & 0.735 \\
     \texttt{Gemini-Pro} & 0.770 & 0.628 \\
     \texttt{Gemini-Flash} & 0.737 & 0.639 \\
    \bottomrule
    \end{tabular}
    \caption{Benchmarking LLMs on INLI. We explore fine-tuning smaller LLMs, as well as few-shot prompting larger LLMs. We see that all models perform worse on implied entailments in INLI than the rest of the dataset, indicating room for improvement in implication understanding capabilities.}
    \label{tab:performance}
\end{table}

\subsection{Benchmarking LLMs on INLI}

How well do modern LLMs perform on INLI? We explore the capabilities of smaller LLMs that can be fine-tuned cheaply (the \texttt{T5} family), as well as larger LLMs that we few-shot prompt (\texttt{GPT-4}, \texttt{Claude-3}, \texttt{Mistral}). Results are in Table~\ref{tab:performance}, and experiment details are in Appendix~\ref{app:experiments-inli-performance}.

Though the fine-tuned LLMs do decently well on INLI overall, the implied entailment accuracy is lower than overall accuracy for all models. Even \texttt{T5-XXL}, which performs the best on INLI overall, has an implied entailment accuracy of 0.885, leaving sufficient room for improvement. Note that the human agreement for implied entailment is 0.94 (see Table~\ref{tab:classwise-annotation}), suggesting a similar upper bound for this task.

To benchmark larger LLMs, we create an 8-shot prompt (see Appendix~\ref{app:prompts}) with examples spanning all hypothesis types and datasets. Though known for high performance on reasoning tasks, these larger LLMs perform worse than the fine-tuned models, and for most LLMs, the discrepancy between overall accuracy and implied entailment accuracy is even more pronounced. 

Note that \texttt{Gemini-Pro} was the model used to build the dataset. Despite this, we see the same pattern -- mediocre overall accuracy and an even lower implied entailment accuracy. These results suggest that models like \texttt{GPT-4} struggle with understanding implication, but fine-tuning on INLI can significantly improve performance. 

\vspace{0.5em}
\noindent \textbf{Takeaway:}  INLI is a challenging dataset for today's LLMs, but training on INLI improves their ability to understand implication.

\begin{table}[t]
\small
    \centering
    \begin{tabular}{lrr}
    \toprule
     & \multicolumn{2}{c}{\textbf{Training Datasets}} \\
    \cline{2-3} \rule{0pt}{1.25em}
    & \textit{Standard NLI} & \textit{3-way INLI} \\
    \midrule
    \textbf{SNLI} & 0.934 & 0.921 \\
    \textbf{MNLI} & 0.916 & 0.914 \\
    \textbf{ANLI} & 0.725 & 0.734 \\
    \textbf{WANLI} & 0.825 & 0.822 \\
    \textbf{3-way INLI} & 0.778 & 0.909 \\
    \bottomrule
    \end{tabular}
    \caption{Models fine-tuned on INLI retain NLI capabilities. We measure accuracy on 4 NLI benchmarks before and after finetuning \texttt{T5-XXL} on INLI.}
    \label{tab:efficacy}
\end{table}

\begin{table*}[t]
    \centering
    \small
    \begin{tabular}{llrlr}
     \toprule
      & \textbf{Training Subset} & \textbf{Accuracy} & \textbf{Generalization Subset} & \textbf{Accuracy} \\
     \midrule
     \multirowcell{2}{\textit{Intra-Domain}\\\textit{Generalization}} & \textsc{NormBank} & 0.917 & \textsc{SocialChem} & 0.795 \\
     & \textsc{SocialChem} & 0.913 & \textsc{NormBank} & 0.850 \\
     \midrule
     \multirowcell{2}{\textit{Cross-Domain}\\\textit{Generalization}} & \textsc{Ludwig, Circa} & 0.965 & \textsc{SocialChem, NormBank} & 0.695 \\
     & \textsc{SocialChem, NormBank} & 0.913 & \textsc{Ludwig, Circa} & 0.796 \\
     \midrule
     \multirowcell{2}{\textit{Cross-Dataset}\\\textit{Generalization}} & \textsc{NormBank, Ludwig, Circa} & 0.945 & \textsc{SocialChem} & 0.804 \\
     & \textsc{SocialChem, Ludwig, Circa} & 0.942 & \textsc{NormBank} & 0.851\\
     \bottomrule
    \end{tabular}
    \caption{INLI supports generalization across datasets and domains. We run various generalization experiments and show how INLI can improve LLM ability to understand implication in new environments.}
    \label{tab:generalization}
\end{table*}

\subsection{INLI and Existing NLI Benchmarks}
We also want to ensure that a model trained to understand implied entailment still maintains efficacy on the standard NLI task. 

To do this, we first fine-tune a \texttt{T5-XXL} model on the four NLI benchmarks listed in Table~\ref{tab:curr-benchmarks}, and assess performance. These preliminary results are in the first column of Table~\ref{tab:efficacy}. We then augment INLI to fit the standard 3-label NLI task, where explicit and implicit entailments are collapsed into the ``entailed'' label. Lastly, we fine-tune the model on augmented INLI, using early stopping to prevent forgetting \cite{DBLP:journals/corr/BaroneHGS17}. Results can be found in the second column of Table~\ref{tab:efficacy}, and experiment details are in Appendix~\ref{app:experiments-inli-efficacy}. 

Encouragingly, we see that performance on INLI increases, while performance on the NLI benchmarks roughly stays the same. Performance on ANLI also increases slightly, suggesting that understanding implication is useful for the challenging examples in ANLI.

\vspace{0.5em}
\noindent \textbf{Takeaway:} Results indicate LLMs trained to infer implication will retain previous reasoning capabilities on other tasks.

\subsection{Generalization Capabilities of \datasetname}

A key goal of INLI is helping LLMs understand implication. So, it is crucial that models trained on INLI have the ability to generalize to unseen domains and datasets.

In order to assess whether INLI is useful for generalizable understanding of implied entailment, we perform a series of generalization experiments to assess how LLMs that learn implication in one environment can apply it to a different environment. We specifically test the following three generalization capabilities:

\begin{enumerate}
\vspace{-0.5em}
\itemsep 0em
    \item \textbf{Intra-domain:} Generalization from one dataset to another in the same domain. We train on \textsc{NormBank} and test generalization to \textsc{SocialChem}, and vice versa.\footnote{We choose \textsc{NormBank} and \textsc{SocialChem} as \textsc{Circa} and \textsc{Ludwig} are quite similar in content/structure, so our situational datasets more accurately assess generalization.}
    \item \textbf{Cross-domain:} Generalization from one domain to another. We train on the situational subset of INLI and test on the conversational subset, and vice versa.
    \item \textbf{Cross-dataset:} Generalization to an unseen dataset. We train on three datasets and test on the fourth, using \textsc{NormBank} and \textsc{SocialChem} as our held-out datasets.
\end{enumerate}

Results are in Table~\ref{tab:generalization} and experiment details are in Appendix~\ref{app:experiments-generalization}. We see impressive generalization results across all experiments -- for instance, a model that has never seen \textsc{NormBank} but is fine-tuned on the other three datasets in INLI has better accuracy on \textsc{NormBank} than \texttt{GPT-4} or \texttt{Claude-3} prompted using examples from all four datasets. 

\vspace{0.5em}
\noindent \textbf{Takeaway:} Models trained on INLI can generalize to understand implication in new domains and environments.

\section{Related Work}

\textbf{Structured Implicatures:} Most past work that focuses on implicatures requires a fixed input structure -- indirect QA \cite{george2020conversational, louis2020d, takayama-etal-2021-direct-direct, damgaard-etal-2021-ill}, scalar implicatures focusing on logical consistency \cite{zheng2021grice, jeretic2020natural}, pairwise entity selection \cite{hosseini2023resolvingindirectreferringexpressions}, etc. However, their fixed input and output structure limits utility; INLI is unstructured and thus more useful for generalizable implicature understanding.

\textbf{Implicature Frames:} A parallel line of work analyzes how the implicit meaning of text changes based on context -- social norms \cite{sap2019social, forbes2020social}, cultural norms \cite{ziems2023normbank, rai2024socialnormscinemacrosscultural}, offensive speech \cite{zhou2023cobra}, etc. We pull from this rich body of work for our situational implicature frames.

\textbf{Implicature Understanding:} Previous work has attempted to measure implicature understanding in LLMs -- via chain-of-thought \cite{kim2023ispopecatholicapplying}, explanation generation \cite{yue2024largelanguagemodelsunderstand}, comparison to humans \cite{qiu-etal-2023-chatgpt}, and intent classification \cite{zhang2024modeltellnegationimplicature}. Findings vary from ``LLMs fail to understand implicatures'' \cite{yue2024large, qiu-etal-2023-chatgpt} to ``LLMs show promise.'' \cite{kim1998high, zhang2024modeltellnegationimplicature}. This discrepancy further motivates the need for an implication-focused benchmark like INLI.

\textbf{Commonsense NLI:} Past commonsense NLI research \cite{rudinger-etal-2020-thinking, zhang-ordinal-commonsense-2017} integrates various types of world knowledge and analyzes downstream likelihood of entailment. These works, along with certain NLI datasets \cite{zellers2019hellaswagmachinereallyfinish, bhagavatula2020abductivecommonsensereasoning, bisk2019piqareasoningphysicalcommonsense}
emphasize understanding physical and temporal constructs for success but do not address the distinction between explicit and implied entailment. \citet{ghosal2021cidercommonsenseinferencedialogue}
touches on implicitness but is limited to causal phrases derived from predefined relations, making it inapplicable to free-form reasoning. INLI focuses on explicit vs. implied information in free-form premises, a boundary not captured by Commonsense NLI. This distinction is highlighted by performance discrepancies (e.g. GPT-4 achieves 95\% on HellaSwag \cite{zellers2019hellaswagmachinereallyfinish}, but 75\% on INLI).

\section{Conclusion}
In this paper, we formalized the implied entailment task, providing an evaluation scheme for implied entailments. We created the Implied NLI (INLI) dataset, a stepping stone to help LLMs reason about and infer implication in text. We demonstrated the utility of INLI in working towards models that understand the subtleties of human language. 

We hope INLI will provide a tool to benchmark LLMs and improve their ability to understand implication. We encourage future researchers to explore how INLI can help LLMs read between the lines and better support real-world communication.

\section*{Limitations}

There are various limitations with the data we augmented to build INLI. Since our objective was to build a dataset that mirrored human communication, we focused on the situational and conversational domains. However, this may lead to a lack of domain diversity poor generalization to formal prose or highly specific text (e.g. medical, legal). 

Since we chose to augment existing data for more reliable implied entailments, any flaws with the original datasets (typos, incorrect labels, etc.) may be propagated up in INLI. Though we had annotators vet a subset of INLI, the entirety of our dataset has not been human-vetted; thus, some examples may be erroneous.

Additionally, the pipeline we used to generate premises, implied entailments, and alternative hypotheses may result in little diversity amongst examples, as generating multiple statements using a single prompt tends to produce similar outputs. Though we attempt to tackle this issue with paraphrasing, our dataset does not wholly capture real-world diversity of language. 

At a higher level, implications are subjective, and personal/cultural context plays a large role in how humans understand implied language. Given a premise in INLI, the implied entailment may follow for some readers, but not all. For instance, people from different cultural backgrounds have a different understanding of norms and behaviors, and might disagree with some examples in INLI.

\section*{Ethics Statement}

The datasets built on in this work, specifically \textsc{NormBank} and \textsc{SocialChem}, focus on social norms/situations and therefore may contain sensitive content and potentially offensive themes. However, we feel is important for LLMs to reason about implications in a diverse variety of settings and situations, so we chose to not filter out these examples.

We also recognize that data produced by large pre-trained language models risks including hallucinations, misinformation, toxicity, and other social harms. As with any generated dataset, INLI is susceptible to these issues. 

Additionally, INLI contains names in the vast majority of its premises, which are randomly sampled from the most popular names in the 2011 U.S. Census \cite{census:2011}. Any premises in INLI are entirely fictitious and bear no resemblance to specific people or events in the real world.

\bibliography{custom}
\bibliographystyle{acl_natbib}

\newpage

\appendix
\section{Experiment Details}
In this section we provide details on the exact experimental setup used in our experiments in order to aid reproducibility of this work.

\subsection{NLI Benchmarks Contain Few Implicatures}
\label{app:experiments-curr-benchmarks}

To fine-tune the implicitness detection model, we use implied entailments and explicit entailments from the training set of INLI. We fine-tune a \texttt{T5-XXL} model using 3 different learning rates: $1e-6, 5e-6, 1e-5$, as they work well for similar previous experiments \cite{hosseini2024synthetic}. We fine-tune for 50,000 steps using a batch size of 32 and ultimately select the checkpoint with the highest validation accuracy upon experiment completion for the results reported in Table~\ref{tab:curr-benchmarks}. 

We then apply this model to the test sets of SNLI, MNLI, ANLI, and WANLI, and report the \% of test set classified as implicit.

\paragraph{Manual validation of classifier.}  Two authors manually annotate a subset of 80 examples to ensure the model is generalizing correctly. We randomly select 10 labeled implied entailments and 10 labeled explicit entailments from each of our four existing benchmarks to ensure a balanced subset. Given a $\langle$premise, hypothesis$\rangle$ pair, annotators are instructed to label the hypothesis as either implicitly entailed or explicitly entailed. We then calculate classifier accuracy by averaging the accuracy with respect to Annotator 1's labels and Annotator 2's label. Accuracies and inter-annotator agreements are given in Table~\ref{tab:classifier-annotation}.

\subsection{NLI Models Rarely Generalize to Implicatures}
\label{app:experiments-curr-models}

We fine-tune a different \texttt{T5-XXL} model for every benchmark in Table~\ref{tab:curr-models}, using the entirety of the training set for each benchmark. We use a learning rate of $5e-6$ and fine-tune for 50,000 steps using a batch size of 32 and select the checkpoint with the highest validation accuracy. We then run the four fine-tuned models on the test set of INLI and report accuracy on the explicit entailment and implied entailment subsets.

\begin{table}[t]
\newcolumntype{R}[1]{>{\raggedleft\let\newline\\\arraybackslash\hspace{0pt}}m{#1}}
\small
    \centering
    \begin{tabular}{lR{2cm}R{2.5cm}}
    \toprule
    \textbf{Dataset} & \textbf{Classifier Accuracy} & \textbf{Agreement (Cohen's Kappa)} \\
    \midrule
    SNLI & 0.925 & 0.875 \\
    MNLI & 0.889 & 0.780 \\
    ANLI & 0.917 & 0.756 \\
    WANLI & 0.917 & 0.667 \\
    \midrule
    \textit{Overall} & 0.920 & 0.768 \\
    \bottomrule
    \end{tabular}
    \caption{Human validation for implicitness detection classifier. We report for what \% of labels the annotators agree with the classifier, and the inter-annotator agreement for each dataset.}
    \label{tab:classifier-annotation}
\end{table}

\subsection{Benchmarking LLMs on INLI}
\label{app:experiments-inli-performance}

To fine-tune the famliy of \texttt{T5} models, we use a learning rate of $5e-5, 1e-4, 5e-4, 1e-3$ for \texttt{XXL, Large, Base, Small} respectively. We fine-tune for 50,000 steps using a batch size of 32 for \texttt{XXL} and 64 for the other models. For each model, we use the checkpoint with the highest validation accuracy. We report results on the full test set of INLI and the implied entailment subset.

To query the larger LLMs, we use an 8-shot prompt (See Appendix~\ref{app:prompts}) and use the open-source API corresponding to each model. For all LLMs, temperature is set to 1. Results in Table~\ref{tab:performance} reflect models queried during August 1st-15th, 2024.

\subsection{INLI and Existing NLI Benchmarks}
\label{app:experiments-inli-efficacy}

We fine-tune a \texttt{T5-XXL} model on the combination of SNLI, MNLI, ANLI, and WANLI training sets. We use 3 different learning rates: $1e-6, 5e-6, 1e-5$. We fine-tune for 50,000 steps using a batch size of 32 and select the checkpoint with the highest validation accuracy upon experiment completion for the results shown in Table~\ref{tab:efficacy}. We report results on the test sets of all 4 NLI benchmarks.

Next, we collapse INLI into a 3-label dataset by mapping our implied entailment and explicit entailment labels to ``entailed.'' Lastly, we fine-tune the same model on INLI for 5,000 steps, implementing early stopping as a strategy to prevent forgetting, a phenomenon observed during transfer learning \cite{DBLP:journals/corr/BaroneHGS17}. We report results once again on the test sets of all 4 NLI benchmarks.

\subsection{Generalization Capabilities of INLI}
\label{app:experiments-generalization}

We fine-tune six \texttt{T5-XXL} models as follows: Using the subset of INLI's training set defined in the first column of Table~\ref{tab:generalization}, we fine-tune the model with a learning rate of $5e-6$ and a batch size of 32. We then report accuracy on the subset of INLI's test set defined in the second column of Table~\ref{tab:generalization}.

\begin{table*}[t]
\centering
\footnotesize
\begin{tabular}{l m{0.35\textwidth} m{0.18\textwidth} m{0.23\textwidth}}
\toprule
\textbf{Dataset} & \textbf{Premise} & \textbf{Implied Entailment} & \textbf{Cognitive Step} \\
\midrule
\textsc{Ludwig} & Diane says, "Would you like to go to a party tonight?" Sophie responds, "I am too tired." & Sophie would prefer not to attend the party this evening. & \textit{Logical Reasoning:} If someone is tired, then they would not attend a social event. \\
\midrule
\textsc{Circa} & Teresa and Alicia are colleagues who are leaving work on a Friday at the same time. Teresa says, “Do you want to hang out later?” Alicia responds, “I could do with a stiff drink.” & Alicia hopes to spend time with Teresa later. & \textit{Conversational Pragmatics:} Alicia is suggesting she and Teresa grab a drink together. \\
\midrule
\textsc{NormBank} & A hush fell over the elegant restaurant as a group of diners, their laughter escalating hysterically, began to fling mashed potatoes and bread rolls at each other, splattering the white tablecloth and surrounding patrons with food. & The other diners were shocked by the inappropriate food fight. & \textit{World Knowledge:} A hush falls over a room when something shocking is happening. \\
\midrule
\textsc{SocialChem} & With a heavy heart, Maya explained to her sister, Chloe, that she wouldn’t be able to donate a kidney, bracing herself for the wave of emotions that were sure to follow. & Maya realized her choice not to donate a kidney would probably devastate Chloe. & \textit{World Knowledge:} When someone asks a family member for an organ donation, they hope the answer is yes. \\
\bottomrule
\end{tabular}
\caption{We specifically select datasets with both conversational and situational implications for a diverse set of cognitive deductions that need to be made. The implications in Circa and Ludwig primarily require conversational pragmatics and logical reasoning to deduce, while the implications in SocialChem and NormBank rely on world knowledge. We show a categorization of cognitive deductions for the examples in Table~\ref{tab:dataset-examples}.}
\label{tab:cognitive-deductions}
\end{table*}

\section{Annotation Study}
\label{app:annotation}

The six annotators chosen for this task are all experts in NLI, and therefore have an established understanding of entailment, neutrality, and contradiction. Annotation instruction mainly focuses on distinguishing implied entailments from explicit entailments. 

Annotators are shown Figure~\ref{fig:plausibility} and given the following evaluation scheme to evaluate premises and hypotheses in INLI:

\begin{figure}[t]
    \centering
    \includegraphics[width=0.9\linewidth]{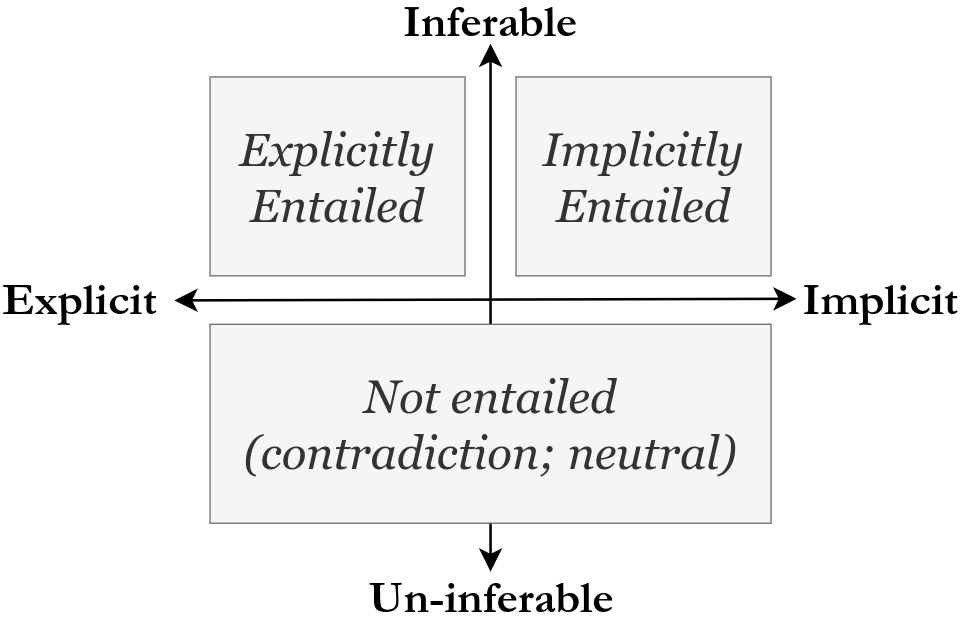}
    \caption{Defining implied entailment. We propose implied entailments should be evaluated via two axes -- inferability and implicitness.}
    \label{fig:plausibility}
\end{figure}

\paragraph{Instructions.} Implied entailments should be defined along two axes: inferability and implicitness. An implied entailment must be \textit{inferable} (i.e. able to be reasonably inferred from the premise) and \textit{implicit} (i.e. requiring a cognitive step based on implicit shared context to infer). The inferability axis measures to what degree a hypothesis can be supported by a premise. For instance, in Figure~\ref{fig:plausibility}, hypotheses that can be inferred, either explicitly or implicitly, are entailed. The implicitness axis measures to what extent a hypothesis requires cognitive steps to infer. Hypotheses that are not directly expressed through lexical syntax or semantics are implicit.

The overall majority agreements (i.e. at least 2 out of 3 annotators agree with the given label) for each class are given in Table~\ref{tab:classwise-annotation}.

\begin{table}[t]
    \centering
    \small
    \begin{tabular}{lr}
    \toprule
    \textbf{Label}	& \% \textbf{Majority Agreement} \\
    \midrule
    Explicit entailment	& 0.96 \\
    Implied entailment & 0.94 \\
    Neutral	& 0.94 \\
    Contradiction & 0.90 \\
    \midrule
    \textit{Overall} & 0.935 \\
    \bottomrule
    \end{tabular}
    \caption{Class-wise annotation results. For each class in INLI, we report the \% of majority annotations that matched the given label.}
    \label{tab:classwise-annotation}
\end{table}

\begin{table*}[t]
\small
    \centering
    \begin{tabular}{lrrrr}
    \toprule
    \multirow{2}{*}{\textbf{Dataset}} & \multicolumn{4}{c}{\textbf{Accuracy}} \\
    \cline{2-5} \rule{0pt}{1.25em}
    & \textbf{Entailed (Implicit / Explicit)} & \textbf{Neutral} & \textbf{Contradiction} & \textbf{Overall} \\
    \midrule
    SNLI \cite{bowman2015large}  & 0.706 & 0.697 & 0.724 & 0.709 \\
    MNLI \cite{williams-etal-2018-broad} &  0.570 & 0.661 & 0.702 & 0.642 \\
    ANLI \cite{nie2020adversarialnlinewbenchmark} & 0.532 & 0.645 & 0.401 & 0.526 \\ \vspace{0.1cm} 
    WANLI \cite{Liu2022WANLIWA} & 0.501 & 0.595 & 0.301 & 0.516 \\ 
    INLI (ours) & 0.684 / 0.727 & 0.496 & 0.475 & 0.595 \\
    \bottomrule
    \end{tabular}
    \caption{Hypothesis-only baseline. We fine-tune a \texttt{T5-XXL} model to label the hypothesis, while entirely excluding the premise. We find the hypothesis-only baseline for our dataset is in line with existing benchmarks, indicating that inference on our dataset does depend on both premise and hypothesis.}
    \label{tab:robustness}
\end{table*}

\section{Developing our Prompting Framework}
\label{app:prompting}
In this section, we will cover the variety of prompting frameworks we experimented with before settling on the framework detailed in Section~\ref{sec:dataset}. We will discuss the intuition behind developing each iteration of the framework and what worked/did not work in regards to the final generated data for each iteration.

All framework iterations below generated hypotheses for \textsc{Ludwig}, \textsc{Circa}, \textsc{NormBank}, \& \textsc{SocialChem101} separately. We also used the same 20 few-shot examples (5 per dataset) across all iterations.

\paragraph{Few-shot examples.} In order to determine high-quality hypotheses based on the premises in the few-shot examples, the authors consulted in-house entailment experts. Together, the authors and their colleagues brainstormed what would be good implied entailments, explicit entailments, neutrals, and contradictions based on the premises, and settled on the final examples shown in Appendix~\ref{app:prompts}.

\subsection{Iteration 1: Separate generation of each hypothesis}
The first attempt of our data generation framework was straightforward and simple. In order to ensure each hypothesis was indeed valid with respect to the premise, we thought it best to create a different prompt for each type of hypothesis (i.e. explicit entailment, neutral, and contradiction). An example in our few-shot prompt contained the example premise, followed by the example hypothesis.

\paragraph{What worked.} The generated hypotheses mimicked the few-shot examples in length, complexity, and sentence structure.
\paragraph{What didn't work.} Many generated hypotheses were not valid with respect to the premise (e.g. a generated ``contradiction'' was actually neutral based on premise).

\subsection{Iteration 2: Adding justifications}

In order to improve the validity of the generated hypotheses, we decided to have the model generate justifications, or explanations of why the hypothesis is indeed contradictory, explicit, etc. Having LLMs provide explanations of their generations \cite{lampinen2022languagemodelslearnexplanations} has been shown to increase accuracy and relevance of generations, especially in the domains of inference and claim extraction \cite{wang2023explainableclaimverificationknowledgegrounded, zeng2024justilm}. An example in our few-shot prompt contained the example premise, followed by the example hypothesis, and a 1-2 sentence justification of why the hypothesis was validly labeled with respect to the premise.

\paragraph{What worked.} The hypotheses were indeed valid with respect to the premise. Incorporating justification significantly increased the correctness of the generated hypotheses.
\paragraph{What didn't work.} Amongst the four hypotheses corresponding to the premise, there was a lot of inconsistency. We noticed neutral hypotheses were much shorter and contradictions nearly always had negation.

\subsection{Iteration 3: All-at-once generation}
We wanted to limit any artifacts in the data that a model might learn that separate hypotheses into classes, irrespective of the premise. To fix the inconsistencies resulting from Iteration 2, we had the model generate all hypotheses at once. An example in our few-shot prompt contained the example premise, followed by the four corresponding hypotheses, and a 1-2 sentence justification for each hypothesis explaining why the hypothesis was validly labeled with respect to the premise.

\paragraph{What worked.} The inconsistencies in sentence length and sentence structure were resolved. 
\paragraph{What didn't work.} We noticed other inconsistencies like word choice (e.g. \textsc{NormBank} implicatures tended to have words like ``inappropriate'' or ``disrespectful'') and theme (e.g. neutral hypotheses tended to be more outlandish).

\subsection{Iteration 5: All-at once generation via modification}
Our last and final iteration aimed to fix the inconsistencies in word choice and theme of the generated hypotheses. To do this, we instructed \texttt{Gemini-Pro} to use the implied entailment as a starting point and modify key words/phrases to transform the implied hypothesis into a contradiction, a neutral, and an explicature.

\paragraph{What worked.} The hypotheses were much more homogeneous. Given we wanted to create a challenging benchmark, the homogeneity made it harder for the model to disentangle hypotheses without using the premise to reason.

\section{Data Robustness}
\label{app:robustness}

Past work from \cite{poliak2018hypothesisbaselinesnaturallanguage} highlights a desired condition for all NLI datasets: \textit{inference should depend on both premise and hypothesis.} One way to measure whether inference is premise-dependent is to fine-tune a model to label the hypotheses, while entirely excluding the premises. 

Table~\ref{tab:robustness} contains the hypothesis-only baseline for our dataset, along with those of other NLI benchmarks. We see our overall hypothesis-only accuracy is in line with other benchmarks, indicating our data does not have many artifacts that may make the task less challenging. 

We do see a higher accuracy in identifying explicit entailments; this is expected, as all of our situational implied entailments focus on violating social norms or reasoning about social rules-of-thumb. However, such social constructs are all implied by nature; therefore, none of our situational explicatures focus on social constructs, while the other hypotheses do.

\section{Prompts}
\label{app:prompts}
Here, we provide the prompt templates used for all of generation tasks described in this paper. For the INLI generation pipeline, prompts vary for each dataset, we provide the full list of examples used here.\footnote{\url{https://github.com/google-deepmind/inli/tree/main/Resources/Prompts}}

\subsection{Implicature Augmentation}
Prompt for Stage 1 of our pipeline to build INLI: Implicature augmentation. This prompt is used to augment implicature frames from \textsc{NormBank} and \textsc{SocialChem} into premises and implicatures. We use a different set of 5 examples for each dataset. 
\vspace{0.5em}

\texttt{\footnotesize
You are an expert in English reading comprehension. You also understand social situations well.
Your task is, given a setting, behavior, and details about a socially taboo situation, construct a passage that depicts the situation being interpreted as taboo in a real-life setting.
Then, extract a logical implicature about why the behavior described in the passage is seen as taboo.
An "implicature" is defined as a statement that can be inferred from the passage, but is not explicitly stated in the passage.
There are many possible ways to do this task. Here are a few examples, with explanations.
\newline \newline
 [----- begin example -----] \newline
<Original Data>
\newline
<Passage> 
\newline
<Implicature> 
\newline
<Justification> 
\newline
[----- end example -----]}

\subsection{Alternative Hypothesis Generation}
Prompt for Stage 2 of our pipeline to build INLI: Alternative hypothesis generation. This prompt is used to generate explicit entailments, neutrals, and contradictions, given the passages and implied entailments generated in Stage 1. We use a different set of 5 examples for each dataset. 

\vspace{0.5em}
\texttt{\footnotesize
You are an expert in English reading comprehension. Given a passage depicting a conversation/social situation, there are four possible types of statements that can be extracted from the passage.
\newline \newline
These four statements are: an implied statement, an explicit statement, a neutral statement, and a contradictory statement. An "implied statement" is defined as a statement that can be inferred from the passage, but is not explicitly stated in the passage.
An "explicit statement" is defined as a statement that is entailed in the passage, and can be proven true using the passage as support.
A "neutral statement" is defined as a statement that is neither entailed nor contradicted by the passage. A neutral statement can be true given the passage, but can also be false given the passage.
A "contradictory statement" is defined as a statement that can be directly refuted and proven false using the passage as support.
\newline \newline
Given a passage and an implied statement, your task is to generate explicit, neutral, and contradictory statements, using the implied statement as a starting point.
You may add, remove, and replace any necessary phrases from the implied statement, so that the generated statements are explicit, neutral, and contradictory with respect to the passage.
\newline \newline
Here are few example passages, the four statements corresponding to the passage, and justifications explaining each of the statements:
\newline \newline
[----- begin example -----] \newline
<Passage>
\newline
<Implied statement> 
\newline
<Explicit statement + justification> 
\newline
<Neutral statement + justification> 
\newline
<Contradictory statement + justification> 
\newline
[----- end example -----]}

\subsection{Benchmarking LLMs on INLI}
Prompt to benchmark \texttt{GPT-4}, \texttt{GPT-4o}, \texttt{Claude-3-Sonnet}, and \texttt{Mistral-Large} on INLI. We select 8 random examples from the training subset of INLI, ensuring that these examples include all statement types and datasets. 

\texttt{\footnotesize
Given a premise and a hypothesis, your task is to label whether the hypothesis is a valid inference from the premise.
Specifically, you will need to assign one of four labels to the hypothesis:
\newline
- Implicature: The hypothesis is a valid inference from the passage, but it is NOT explicitly stated in the passage.\newline
- Explicature: The hypothesis is a valid inference from the passage, and it is explicitly stated in the passage.\newline
- Contradiction: The hypothesis is NOT a valid inference from the passage, and is contradicted by the passage.\newline
- Neutral: The hypothesis can neither be supported nor contradicted by the passage.
 \newline \newline
Here are some examples:
\newline \newline
[----- begin examples -----] \newline
Premise:  Ashley felt a pang of guilt as she accepted a date with Jake, knowing her friend Lisa had briefly dated him a few months prior. \newline
Hypothesis:  Ashley is aware that dating Jake might make Lisa, her friend, upset. \newline
Label:  Implicature \newline \newline
Premise:  Diana wants to know what activities Paula likes to do during weekends. Diana says, "Do you like local theater?" Paula responds, "Only musicals." \newline
Hypothesis:  Paula likes local theater, but only musicals. \newline
Label:  Explicature \newline \newline
Premise:  As the professor dismissed the class, a classmate approached Daniel, but Daniel, eager to get to his next class, rushed past with a quick "excuse me" and a fleeting wave. \newline
Hypothesis:  Daniel thinks that socializing during class is inappropriate. \newline
Label:  Neutral \newline \newline
Premise:  Bryon has just travelled from a different city to meet Rodney. Rodney says, "Do you want some time on your own to explore?" Bryon responds, "I'm happy for company." \newline
Hypothesis:  Bryon desires some solitary time for exploration. \newline
Label:  Contradiction \newline \newline
Premise:  Corrine says, "You're gonna be seeing a lot of me. You're sure you don't mind?" Russel responds, "It's good what you're doing." \newline
Hypothesis:  Russel is fine with seeing a lot of Corrine. \newline
Label:  Implicature \newline \newline
Premise:  A bead of sweat trickled down Edgar's temple as he finished mopping the floor by the ellipticals. He paused, catching his breath just as Mr. Robinson, the gym's quiet janitor, awkwardly swung his leg over the machine and began slowly pedaling. \newline
Hypothesis:  Edgar observed Mr. Robinson, the gym custodian, clumsily mount the exercise bike and start pedaling at a leisurely pace. \newline
Label:  Explicature \newline \newline
Premise:  Frieda says, "Have you had a chance to wear your new shirt yet?" Genevieve responds, "I have been trying to exchange it for a larger size." \newline
Hypothesis:  Genevieve likes to wear shirts that are pre-owned. \newline
Label:  Neutral \newline \newline
Premise:  A group of teenagers giggled as they snapped photos of a family struggling to handle their overexcited toddler, their laughter echoing through the primate house.  Dr. Lewis, the zoo's veterinarian, frowned, shaking his head as he walked past. \newline
Hypothesis:  Dr. Lewis smiled as the teenagers photographed unsuspecting strangers at the zoo. \newline
Label:  Contradiction \newline
[----- end examples -----] \newline \newline
[your task] \newline
Given a premise and a hypothesis, your task is to label the hypothesis as one of the four labels: Implicature, Explicature, Contradiction, or Neutral.
Your response should be only one word, the name of the label. \newline
Premise:\{\} \newline
Hypothesis:\{\} \newline
Label:  \newline
}

\end{document}